\documentclass[11pt,a4paper]{article}
\usepackage{times,latexsym}
\usepackage{url}
\usepackage[T1]{fontenc}
\usepackage{graphicx}
\usepackage{algorithm}
\usepackage[noend]{algpseudocode}
\usepackage{siunitx}
\usepackage{url}
\usepackage{pifont}
\usepackage{xcolor}
\usepackage{tablefootnote}
\usepackage{comment}
\usepackage{multirow, makecell}
\usepackage{enumitem}
\usepackage{booktabs}
\usepackage{needspace}
\usepackage{amsmath}
\usepackage[acceptedWithA]{tacl2021v1}

\usepackage{xspace,mfirstuc,tabulary}

\newif\iftaclinstructions
\taclinstructionsfalse 
\iftaclinstructions
\renewcommand{\confidential}{}
\renewcommand{\anonsubtext}{(No author info supplied here, for consistency with
TACL-submission anonymization requirements)}
\newcommand{\instr}
\fi

\iftaclpubformat 

\else

\fi

\title{Efficient ASR Training with Conversations that Never Happened}

\author{
  Máté Gedeon$^{\diamond\dagger}$
  \and
  Péter Mihajlik$^{\diamond\ddagger}$
  \\
  \ \\
  $^{\diamond}$Dept. of Telecommunications and Artificial Intelligence, \\ Budapest University of Technology and Economics,
  Budapest, Hungary
  \ \\
  $^{\dagger}$SpeechTex Ltd.,
  Budapest, Hungary
  \ \\
  $^{\ddagger}$ELTE Research Centre for Linguistics , 
  Budapest, Hungary
  \  \\
  \texttt{gedeonm@edu.bme.hu,mihajlik@tmit.bme.hu}
}

\date{}

\begin{document}
\maketitle
\begin{abstract}
  Conversational ASR for lower-resource languages and niche domains is limited by the scarcity of domain-matched multi-speaker training data. We propose an augmentation pipeline that generates scenario-level dialogues with participant metadata, maps speaker attributes to TTS voice profiles, and assembles synthesized utterances into speaker-aware simulated conversations. We evaluated five LLM families under single-generator, fixed-budget mixture, and scale-up settings using the same FastConformer-Large training recipe for each one. We ran comprehensive evaluations on the Hungarian BEA-Dialogue benchmark corpus, with the method itself being applicable to any language given the resources for each component. The results show that synthetic conversations consistently improve speech recognition performance, but generator choice and data composition strongly affect the gains. Our largest training configuration, using only 67 hours of real conversations and 636 hours of simulated data, achieves better performance on the evaluation benchmark than a zero-shot model trained on 2700 hours of Hungarian speech. These findings indicate that LLM-generated conversational data synthesized with TTS is a practical complement to real conversational corpora for speech model training.
\end{abstract}

\section{Introduction}
Automatic speech recognition (ASR) has improved substantially with deep acoustic modeling and large-scale self-supervised pretraining \citep{Hinton2012,Chan2016,Baevski2020,Radford2023}. However, these gains remain strongly data-dependent: systems trained for high-resource languages benefit from massive transcribed corpora, while low-resource languages and niche domains still face severe data bottlenecks \citep{Besacier2014}. For conversational ASR, the bottleneck is particularly acute because available corpora underrepresent the speaker diversity, discourse phenomena, and topic variation seen in real-world interactions.

Data augmentation partly alleviates this problem. Signal-level methods such as speed/noise perturbation and masking improve robustness \citep{Ko2015,Park2019}, and neural text-to-speech (TTS) enables scalable speech generation from text \citep{Wang2017Tacotron,Shen2018Tacotron2}. Yet existing pipelines usually start from fixed text resources and provide limited control over who is speaking, in what context they are speaking, and how speaker characteristics influence the resulting audio. Consequently, they do not fully capture the interaction between linguistic content, speaker attributes, and multi-party conversational structure.

In parallel, large language models (LLMs) have shown strong ability to generate coherent, context-rich text and dialogue \citep{Brown2020,Ouyang2022}. This creates an opportunity to move beyond transcript-level augmentation toward scenario-based conversational synthesis. The central question is whether LLMs can be used to generate structured conversations that are not only fluent, but also useful for downstream ASR training when coupled with TTS synthesis.

We introduce a unified pipeline for synthetic conversational data generation and evaluate it on Hungarian conversational ASR. The pipeline first prompts an LLM to generate a scenario, participant metadata, and a structured dialogue. It then maps speaker attributes such as age and gender to TTS voice profiles and synthesizes each turn. Finally, it constructs a multi-speaker conversational waveform using speaker-aware simulation \cite{gedeon_icassp}, including pauses and overlap patterns. Although our experiments focus on Hungarian BEA-Dialogue \cite{Gedeon2026BEA}, the pipeline is portable to any language for which a usable TTS system and speaker-reference bank are available.

In our evaluations, we compare five contemporary LLM families (GPT \cite{gpt5}, Claude Haiku \cite{claude}, Gemini \cite{gemini}, Grok\footnote{\url{https://x.ai/news/grok-4}}, Qwen \cite{qwen3}), first measuring each generator in isolation to quantify the impact of individual generators on downstream ASR performance, then testing fixed-budget generator mixtures to assess whether different generators provide complementary benefits, and finally scaling the strongest generator combinations by adding all available synthetic data to evaluate the effect of increasing synthetic data volume on performance. We show that LLM-generated conversations consistently improve training results while also revealing important limitations.

Our main contributions are:
\begin{itemize}
\item A unified framework that combines LLM scenario generation, TTS synthesis, and multi-speaker conversation construction for ASR augmentation.
\item A metadata-conditioned voice-selection procedure that maps generated speaker age and gender attributes to the most suitable reference profiles.
\item A comparative evaluation across multiple contemporary LLM families (GPT, Grok, Gemini, Qwen, and Claude Haiku) under a shared augmentation protocol.
\item Empirical evidence that LLM-driven, speaker-aware synthetic conversations improve Hungarian conversational ASR performance over both non-augmented and prior augmentation baselines.
\end{itemize}

The remainder of this paper is organized as follows. Section 2 reviews related work. Section 3 presents the proposed methodology, while Section 4 describes the experimental framework and setup. Section 5 reports and discusses the results. Finally, Section 6 concludes the paper.

\section{Related Work}
ASR research has progressed from deep acoustic modeling to end-to-end neural and large-scale pretrained systems. Deep neural acoustic models replaced earlier feature-engineered pipelines in many settings \citep{Hinton2012}, while connectionist temporal classification and encoder--decoder approaches made it possible to train recognizers directly from unsegmented speech--text pairs \citep{Graves2006CTC,Chan2016,Amodei2016DeepSpeech2}. More recent architectures and pretraining methods, including Conformer models \citep{Gulati2020Conformer}, wav2vec 2.0 \citep{Baevski2020}, HuBERT \citep{Hsu2021HuBERT}, and weakly supervised Whisper-style training \cite{Radford2023}, have further improved robustness and transfer. Nevertheless, the application of these advances remain strongly data-dependent: lower-resource languages still suffer from limited transcribed speech, domain mismatch, and sparse coverage of conversational phenomena \citep{Besacier2014}. This motivates augmentation methods that can add not only acoustic variation, but also conversational diversity.

Data augmentation for ASR has traditionally focused on transforming existing audio. Speed perturbation and related acoustic transformations increase robustness while preserving the transcript \citep{Ko2015}, and feature-space masking methods such as SpecAugment improve generalization without requiring additional labels \citep{Park2019}. These approaches are effective and widely used, but they cannot introduce new lexical content, speaker roles, discourse structures, or topic variation. 

Synthetic speech generation offers a complementary route: neural TTS systems such as WaveNet \citep{Oord2016WaveNet}, Tacotron \citep{Wang2017Tacotron, Shen2018Tacotron2}, FastSpeech \citep{Ren2019FastSpeech}, and multilingual zero-shot TTS models \citep{xtts} make it possible to synthesize speech from arbitrary text. For conversational ASR, however, utterance-level TTS is not sufficient by itself, because useful training examples must also represent speaker changes, pauses, interruptions, overlaps, and multi-turn context. In English, where conversational TTS models are readily available \cite{Pheme, Guo2021, FireRedTTS}, these models are a natural choice for dialogue synthesis, consistent with the approach adopted by \citet{Cornell2024}. However, such models remain unavailable or insufficiently developed for most languages, necessitating the use of alternative methods for generating conversational speech.

Language modeling solves the content-generation side of this problem. Statistical language modeling has long been central to speech and language processing \citep{GOODMAN2001403,Jurafsky2026}, and Transformer-based pretraining substantially expanded the ability of models to represent and generate fluent text \citep{Vaswani2017,Devlin2019BERT}. Large language models and instruction-tuned systems can generate coherent scenarios, speaker profiles, and multi-turn dialogues \citep{Brown2020,Ouyang2022}. In ASR augmentation, this capability is attractive because it can create new topics and conversational situations rather than merely perturbing existing transcripts. At the same time, generated text is only useful if it can be converted into speech that helps during training; the present work therefore treats text generation, speech synthesis, and conversation simulation as a single pipeline rather than as independent steps.

Conversational speech also differs from read or isolated utterance data because it has an interactional organization. Seminal work on turn-taking shows that conversations are governed by systematic patterns of speaker changes, silence, and overlap \citep{Sacks1974}. Standard conversational and meeting corpora such as Switchboard \citep{Godfrey1992Switchboard}, the ICSI Meeting Corpus \citep{Janin2003ICSI}, and AMI \citep{Carletta2005AMI} have shaped research on spontaneous multi-speaker recognition and interaction modeling. In Hungarian, BEA-Dialogue and BEA-Large \citep{Gedeon2026BEA} provide domain-relevant resources for studying conversational ASR, but the amount of available labeled conversational data remains limited. These corpora highlight the same issue addressed in this paper: conversational ASR requires training data that captures both acoustic variability and interactional structure.

Simulated conversation methods provide a practical way to create such structure when fully natural conversational recordings are scarce. Earlier work commonly focused on concatenating isolated utterances, modeling speaker turns with sampled silence durations, or artificially introducing overlap patterns to approximate spontaneous interaction \cite{Landini2022, Yamashita2022Naturalness, Fujita2019}. More recent approaches have incorporated speaker-aware simulation strategies to better reproduce the temporal structure of natural dialogue \cite{gedeon_icassp,Gedeon2026SASC}, including turn-taking behavior, interruptions, and overlapping speech. In the Hungarian domain, experiments on the BEA-Dialogue corpus showed that adding synthetic conversations by combining BEA-Large utterances with sampled pauses, speaker changes, and overlaps improves conversational ASR performance \citep{Gedeon2026SASC}. Our work builds upon this line of research by replacing fixed utterance inventories with dynamically generated LLM-based scenarios and dialogues, while preserving explicit speaker metadata and conversation-level simulation. The resulting pipeline integrates large language models (LLMs), text-to-speech (TTS) systems, and conversation simulation.

\section{Methodology}
We propose a three-stage pipeline for creating synthetic multi-speaker conversational audio for ASR augmentation: (i) LLM-based scenario and dialogue generation, (ii) metadata-conditioned TTS synthesis, and (iii) speaker-aware conversation simulation. Although we instantiate the pipeline for Hungarian, the stages can be transferred to other settings when comparable language resources are available.

\subsection{Problem setup and notation}
Let $\mathcal{D}_{\mathrm{real}}=\{(x_i,y_i)\}_{i=1}^{N}$ denote the original ASR training set, where $x_i$ is an audio segment and $y_i$ is its transcript. Our goal is to construct a synthetic conversational set $\mathcal{D}_{\mathrm{syn}}=\{(\tilde{x}_k,\tilde{y}_k)\}_{k=1}^{K}$ and train on the union
\begin{equation}
\mathcal{D}_{\mathrm{train}} = \mathcal{D}_{\mathrm{real}} \cup \mathcal{D}_{\mathrm{syn}}.
\end{equation}

We optimize the average loss over the combined real and synthetic data.

\subsection{Stage I: LLM-based scenario and dialogue generation}
Following the strong text generation capabilities of modern LLMs \citep{Brown2020,Ouyang2022}, we use a two-step prompting protocol.

\paragraph{Scenario generation.}
For each intended synthetic conversation $k\in\{1,\ldots,K\}$, we generate a scenario
\begin{equation}
c_k = (t_k, M_k), \qquad M_k=\{m_{k,1},\ldots,m_{k,S_k}\},
\end{equation}
where $t_k$ is the topic and $m_{k,j}$ is metadata for speaker $j$. In our setting, each metadata tuple is
\begin{equation}
m_{k,j} = (a_{k,j}, g_{k,j}, o_{k,j}, r_{k,j}),
\end{equation}
with age $a$, gender $g$, occupation $o$, and conversational role $r$.

To improve diversity and realism, the prompts encourage topic diversity across samples, a preference for autobiographical or experiential discussion, avoidance of overly broad or overly niche topics, and realistic variation in ages and occupations.

\paragraph{Dialogue generation.}
Conditioned on $c_k$, we generate a turn-level dialogue
\begin{equation}
U_k = \big((z_{k,1},u_{k,1}),\ldots,(z_{k,T_k},u_{k,T_k})\big),
\end{equation}
where $u_{k,t}$ is the text at turn $t$ and $z_{k,t}\in\{1,\ldots,S_k\}$ is the active speaker index. We use a fixed exemplar dialogue in the prompt to stabilize style and expected turn structure. Generated outputs are filtered by basic quality checks (valid speaker tags, non-empty turns, correct formatting).

\subsection{Stage II: Metadata-conditioned TTS synthesis}
Each utterance $u_{k,t}$ is synthesized with a speaker profile selected from a reference pool. We use a neural TTS model (xTTS-v2) and generate speech sentence-by-sentence, leveraging its voice cloning capabilities.

Let $\mathcal{R}$ denote the reference bank, where each sample $r\in\mathcal{R}$ has demographic annotations $(a(r), g(r))$. For speaker $j$ in scenario $k$, we select
\begin{equation}
\rho^*_{k,j} = \arg\min_{r\in\mathcal{R}:\, g(r)=g_{k,j}}\, |a(r)-a_{k,j}|,
\end{equation}
i.e. the closest profile in age among references with matching gender. We then synthesize
\begin{equation}
\hat{x}_{k,t} = G\big(u_{k,t}, \rho^*_{k,z_{k,t}}\big),
\end{equation}
where $G(\cdot)$ is the TTS model. This mapping provides explicit control over speaker identity while preserving the linguistic diversity of LLM generation. Profile selection can be performed with or without replacement, depending on the size of the reference bank relative to the number of generated conversations.

\subsection{Stage III: Speaker-aware conversation simulation}
The synthesized turns $\{\hat{x}_{k,t}\}_{t=1}^{T_k}$ are converted into a single conversational waveform using the speaker-aware simulated conversation methodology \cite{gedeon_icassp}. Let $d_{k,t}$ be the duration of utterance $t$. We define start times recursively as
\begin{equation}
\tau_{k,1}=0.
\end{equation}
\begin{equation}
\tau_{k,t}=\tau_{k,t-1}+d_{k,t-1}+\Delta_{k,t} \quad (t>1).
\end{equation}
where $\Delta_{k,t}$ is a sampled inter-turn offset. If $z_{k,t}=z_{k,t-1}$, then $\Delta_{k,t}\sim P_{\mathrm{same}}$; otherwise $\Delta_{k,t}\sim P_{\mathrm{switch}}$. Negative values allow overlap, while positive values model pauses.

The final mixed conversation is
\begin{equation}
\tilde{x}_k(\tau) = \sum_{t=1}^{T_k} \hat{x}_{k,t}(\tau-\tau_{k,t}),
\end{equation}
with transcript $\tilde{y}_k$ formed by the ordered turns and their timestamps. The purpose of this stage is to produce training examples that better match conversational ASR conditions (turn transitions, variable pauses, and occasional overlap) than isolated-sentence synthesis, without relying on conversational TTS models, which are not available for most languages.

\subsection{Output of the pipeline}
The pipeline outputs timestamped multi-speaker conversations and aligned transcripts suitable for end-to-end ASR training. In the experimental section, we vary the number of generated conversations and the LLM used in Stage~I to quantify gains from synthetic augmentation and sensitivity to generator choice.

\section{Experiments}
We design the experiments to answer three research questions: (i) whether the proposed synthetic conversations improve conversational ASR over training on real data alone, (ii) how strongly downstream performance depends on the LLM used to generate the conversations, and (iii) whether mixtures of generators yield complementary gains. To this end, we keep the ASR architecture and training recipe fixed and vary only the composition and amount of synthetic data.

\subsection{Synthetic data generation}
\subsubsection{LLM back-ends}
We generate scenarios and dialogues using five LLM back-ends accessed via hosted APIs. In preliminary trials, these systems provided more reliable Hungarian generations than the open-weight alternatives available to us, while remaining inexpensive enough for large-batch data generation. All models are queried with the same prompt templates and controlled generation settings, and we use batch inference when available to reduce cost. 

Despite using identical prompts, the generated datasets differ in total duration due to model-specific generation behavior, primarily variations in average response length. Since any post-processing aimed at equalizing dataset sizes could compromise the fairness of the comparison, the datasets were used without further modification. As a result, the amount of training data is not identical across experiments, and this factor is considered when analyzing the results.

Table~\ref{tab:llm_generation} summarizes the generators used in the study. For ease of reference, in the rest of the paper we refer to each system only by its LLM family name, not by the specific model.  

\begin{table}[t]
\centering
\small
\begin{tabular}{lrr}
\hline
\textbf{LLM} & \textbf{Cost (USD)} & \textbf{Audio (h)} \\
\hline
GPT-5.4 mini\tablefootnote{gpt-5\_4-mini-2026-03-17} & 5.81 & 146 \\
Claude Haiku 4.5 & 8.70 & 127 \\
Gemini 3.5 Flash & 4.50 & 85 \\
Grok 4.1 (non-reasoning) & 0.55 & 81 \\
Qwen3-235B-A22B & 2.62 & 73 \\
\hline
\end{tabular}
\caption{LLM back-ends used for synthetic data generation. Cost denotes the approximate API expenditure for the full generation run after batch discounting, and Audio denotes the duration of the final synthetic speech dataset after TTS and conversation simulation.}
\label{tab:llm_generation}
\end{table}

\subsubsection{TTS reference bank}
For speech synthesis, we use xTTS-v2\footnote{\url{https://huggingface.co/coqui/XTTS-v2}}~\cite{xtts} along with a reference bank from speech segments in BEA-Large \citep{Gedeon2026BEA}. Candidate reference utterances are filtered to durations between 3.5 and 15 seconds, which is sufficient for stable voice cloning while avoiding excessively long prompts. We further restrict the bank to modules containing spontaneous speech, because the target domain is conversational ASR rather than read speech. We also exclude speakers appearing in the BEA-Dialogue development or evaluation sets. After filtering, the bank contains 287 speakers with sufficient age and gender coverage for metadata-matched reference selection.

\subsubsection{Conversation simulation}
To convert utterance-level TTS outputs into conversational training data, we adopt the speaker-aware simulated conversation framework of \citet{Gedeon2026SASC}. We reproduce the best-performing configuration reported in that work using conversational statistics extracted from BEA-Dialogue \citep{Gedeon2026BEA}. In particular, we keep the published kernel bandwidths and simulation parameters fixed, thereby isolating the effect of the proposed text generation stage from changes to the timing model. 

\subsection{Evaluation protocol}
We augment the BEA-Dialogue training split with synthetic conversations and evaluate exclusively on the held-out BEA-Dialogue eval split \citep{Gedeon2026BEA}. BEA-Dialogue is a multi-speaker conversational speech corpus derived from the BEA spontaneous speech database \cite{bea2014} and designed specifically for dialogue-oriented automatic speech recognition experiments. The corpus consists of 11,662 dialogue segments with an average duration of approximately 26 seconds, totaling 84.9 hours of speech. Each segment contains naturally occurring interactions between target speakers, experiment leaders, and discourse partners. The dataset includes both speaker turns and overlapping speech, thereby reflecting some of the challenges encountered in real-world conversational ASR. To enable rigorous evaluation, the training, development, and evaluation partitions are fully disjoint with respect to all speaker roles, preventing speaker leakage across subsets.

We use no synthetic data for evaluation, relying only on real conversations from the BEA-Dialogue corpus. The primary evaluation metrics are concatenated minimum-permutation word error rate and character error rate (cpWER and cpCER) \cite{chime}. All systems share the same ASR architecture, optimization setup, and decoding configuration; consequently, differences in performance can be attributed to the composition of the augmentation data. Model training was carried out using the NVIDIA NeMo framework \cite{nemo}. To ensure comparability between our experiments and prior work, we adhered to the same experimental setup and fine-tuned an English FastConformer~\cite{fastconformer} Large CTC model~\footnote{\url{https://huggingface.co/nvidia/stt_en_fastconformer_ctc_large}} across all configurations. All experiments were conducted on a single \textit{NVIDIA RTX 5000 Ada Generation} GPU with 32 GB of VRAM using a batch size of 16, a learning rate of $5 \times 10^{-4}$, and a cosine annealing learning-rate scheduler.

\begin{figure*}[t]
\centering
\includegraphics[width=0.95\linewidth]{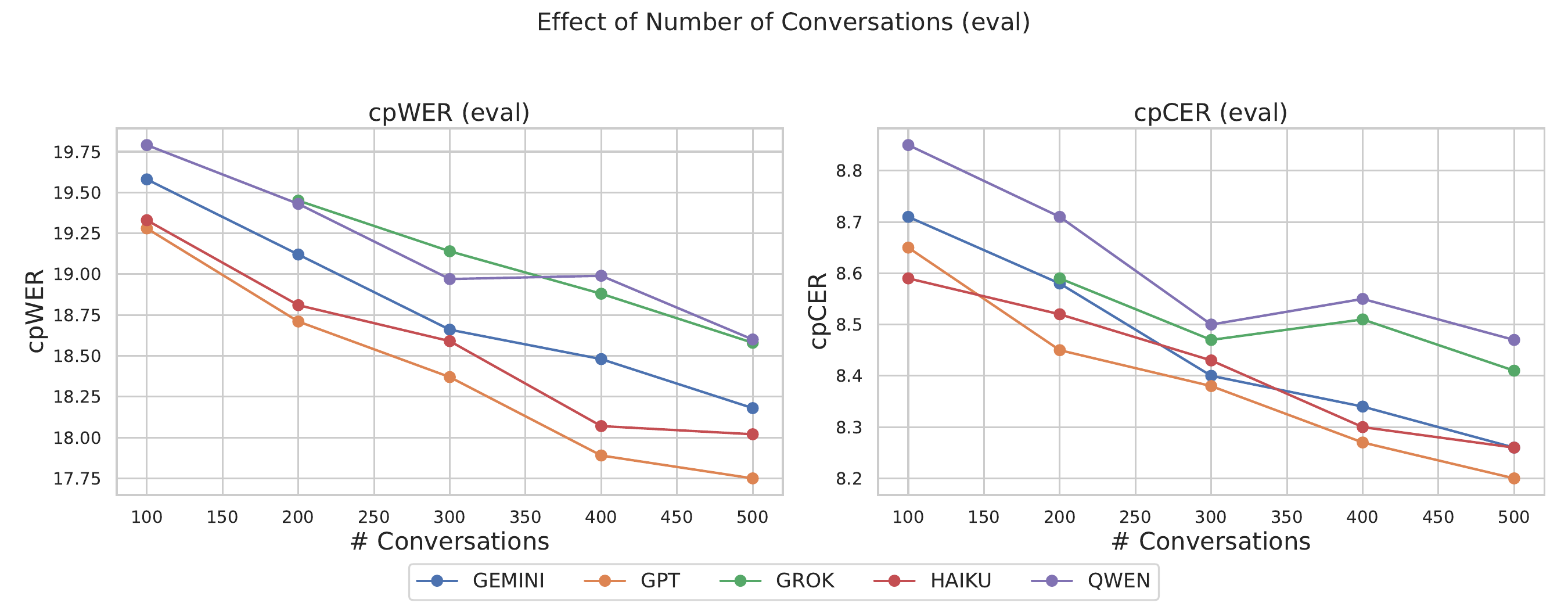}
\caption{Ablation (scale-down) comparison for the single-generator LLM setup.}
\label{fig:ablation_results}
\end{figure*}

\paragraph{Single-generator scaling.}
To measure the utility of each LLM in isolation, we create five synthetic subsets per generator with $K \in \{100, 200, 300, 400, 500\}$ conversations. This yields 25 single-generator conditions and directly measures both the effect of generator identity and the effect of synthetic-data scale.

\paragraph{Mixture study.}
We next test whether different generators contribute complementary conversational patterns. For this study, we fix the total synthetic budget to 500 conversations and form uniform mixtures across the participating LLMs: first pairwise mixtures, then three-way mixtures, then four-way mixtures, and finally a five-way mixture. We progressively add the generators to assess whether diversity across LLM families improves downstream recognition beyond the strongest single-generator setup.

\paragraph{Scale-up study.}
Finally, for the best mixture at each cardinality, we remove the budget restriction and use the total amount of synthetic data available for each component. This experiment tests whether gains persist when the synthetic-to-real ratio increases, or whether performance saturates once the synthetic pool becomes sufficiently large.

\subsection{Baselines}
We compare the proposed systems against an extensive set of baselines designed to separate the effect of conversational simulation from the effect of LLM-driven content generation. The first two baselines are from the original BEA-Dialogue paper \cite{Gedeon2026BEA}.
\begin{itemize}
\item \textbf{Whisper (zero-shot)}: zero-shot results of Whisper-large-v3\footnote{\url{https://huggingface.co/openai/whisper-large-v3}}.
\item \textbf{Real-only}: a FastConformer-Large model fine-tuned only on the BEA-Dialogue training split.
\item \textbf{SASC augmentation}: the same backbone trained on BEA-Dialogue augmented with speaker-aware simulated conversations built from BEA-Large utterances, using the baseline result reported by \citet{Gedeon2026SASC}.
\item \textbf{Hungarian monolingual (zero-shot)}: a FastConformer-Large model created by \citet{Dobsinszki2025} trained on 2700 hours of labeled Hungarian speech. The BEA-Dialogue corpus was not part of the training set, although the training data included substantial broadcast news and broadcast conversations. The BEA-Dialogue zero-shot result is published by \citet{gedeon2026specom}.
\end{itemize}

\begin{table*}[t]
\centering
\small
\begin{tabular*}{\textwidth}{@{\extracolsep{\fill}} l l c c c}
\toprule
\multirow{2}{*}{\textbf{Mix size}} &
\multirow{2}{*}{\textbf{Best subset}} &
\multirow{2}{*}{\textbf{cpCER}} &
\multirow{2}{*}{\textbf{cpWER}} &
\textbf{$\Delta$ cpWER} \\
&&&& vs. best 1-mix \\
\midrule
1-mix & GPT & 8.20 & 17.75 & 0.00 \\
2-mix & \textbf{GPT + Haiku} & \textbf{8.19} & \textbf{17.56} & \textbf{-0.19} \\
3-mix & GPT + Haiku + Qwen & 8.22 & 17.87 & +0.12 \\
4-mix & GPT + Haiku + Qwen + Grok & 8.29 & 18.19 & +0.44 \\
5-mix & GPT + Haiku + Qwen + Grok + Gemini & 8.35 & 18.27 & +0.52 \\
\bottomrule
\end{tabular*}
\caption{Best-performing generator subset at each mixture cardinality (BEA-Dialogue eval set). The last column reports the absolute change in eval cpWER relative to the best single-generator configuration; negative values indicate improvement.}
\label{tab:mixture_summary}
\end{table*}

\section{Results}
\label{sec:results}
We evaluate whether synthetic conversations generated from LLM scenarios improve Hungarian conversational ASR. All systems use the same FastConformer-Large training recipe and are initialized from the same English checkpoint\footnote{\url{https://huggingface.co/nvidia/stt_en_fastconformer_ctc_large}}, so differences among them primarily reflect the synthetic data used for training. We report held-out BEA-Dialogue eval-set cpCER and cpWER; lower values indicate better recognition accuracy.

\subsection{Single-generator setup}
Table~\ref{tab:single_generator_best} summarizes the best result for each LLM family when a single generator is used. All five generators obtain their best reported operating point at the largest evaluated scale, $K=500$, which indicates that the synthetic data remains useful throughout the tested range. GPT-5.4 mini is the strongest standalone generator, reaching 8.20 cpCER and 17.75 cpWER. Haiku 4.5 and Gemini 3.5 Flash match each other in cpCER (8.26), but their differing cpWER suggests more word-level substitution, insertion, or deletion errors for Gemini. Grok 4.1 and Qwen3-235B-A22B also improve over BEA-Dialogue-only training (Table~\ref{tab:scaleup_summary}), but both remain above 18.5\% cpWER, leaving a clear gap compared to GPT, Haiku, and Gemini. Notably, the model ranking  coincides with generated dataset size, although dataset size is not the only factor. For example, Gemini exceeds Grok by only 4 hours of generated audio, but the difference in cpCER and cpWER is much larger between Gemini and Grok than between Grok and Qwen or between Gemini and Haiku.

\begin{table}[t]
\centering
\small
\resizebox{0.48\textwidth}{!}{%
\begin{tabular}{lccc}
\toprule
\textbf{LLM} & \textbf{Best $K$} & \textbf{cpCER} & \textbf{cpWER} \\
\midrule
GPT-5.4 mini & 500 & \textbf{8.20} & \textbf{17.75} \\
Claude Haiku & 500 & 8.26 & 18.02 \\
Gemini 3.5 Flash & 500 & 8.26 & 18.18 \\
Grok 4.1 (non-reasoning) & 500 & 8.41 & 18.58 \\
Qwen3-235B-A22B & 500 & 8.47 & 18.60 \\
\bottomrule
\end{tabular}%
}
\caption{Best single-generator configuration of each LLM family among the available scales. All metrics are computed on the held-out BEA-Dialogue eval set.}
\label{tab:single_generator_best}
\end{table}

Figure~\ref{fig:ablation_results} shows the full scaling behavior. The dominant trend is that adding more synthetic conversations improves recognition, especially when moving from the smallest budgets to $K=500$. The curves are not perfectly monotonic for every generator and metric, however: Grok and Qwen change relative order around the intermediate scales in cpWER, and both Grok and Qwen show a weaker point at $K=400$ than at $K=300$ in cpCER. These fluctuations suggest that scale alone is not the only driver of performance; the linguistic and acoustic properties of the generated conversations also affect how well the augmented data transfers to ASR training. This figure also illustrates the behavior of GPT and Haiku when their generated datasets are closer in duration (hours) to those produced by other models that generated shorter dialogues. Notably, GPT with $K=300$ performs slightly worse than Gemini with $K=500$, but outperforms both Qwen and Grok at $K=500$. It is important to note, however, that in this comparison Qwen and Grok benefit from having generated a substantially larger number of conversations and from exposure to a greater diversity of TTS voices.

\subsection{Generator mixtures}
\begin{figure*}[ht]
\centering
\includegraphics[width=0.9\linewidth]{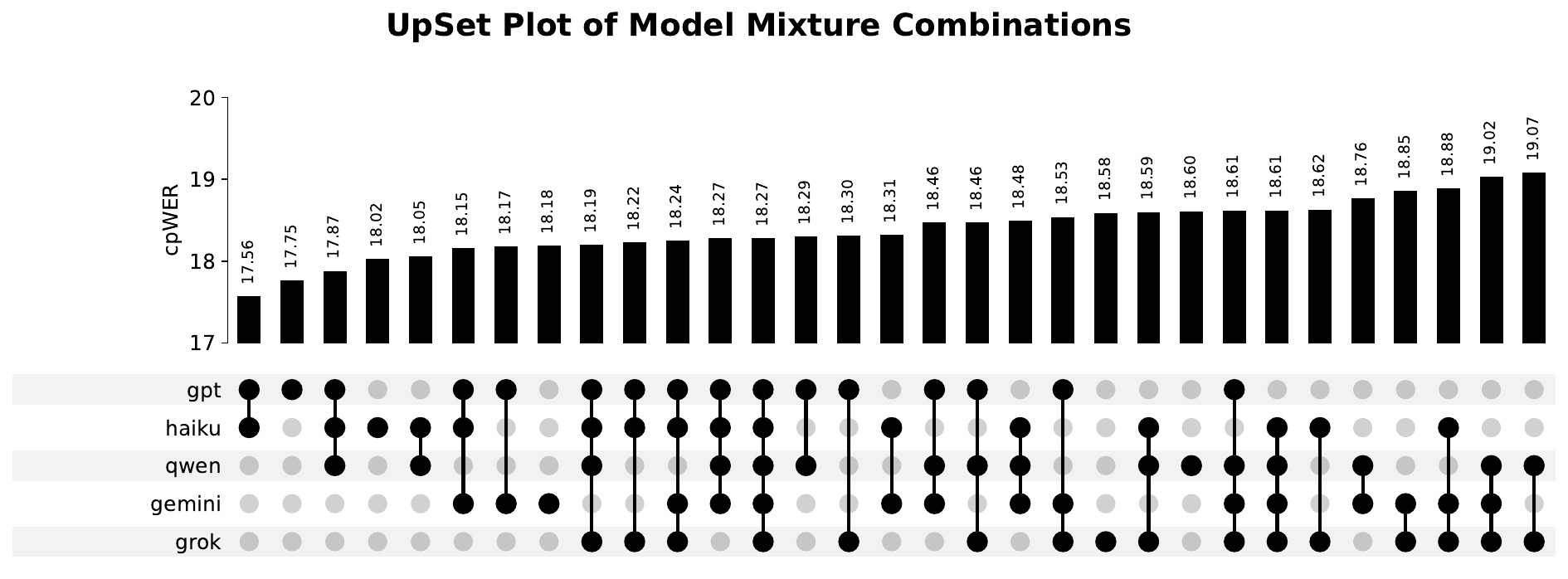}
\caption{Subset-level cpWER (BEA-Dialogue eval) across generator mixtures.}
\label{fig:mixture_upset}
\end{figure*}

\begin{figure*}[t]
    \centering

    \begin{minipage}{0.49\textwidth}
        \centering
        \includegraphics[width=\linewidth]{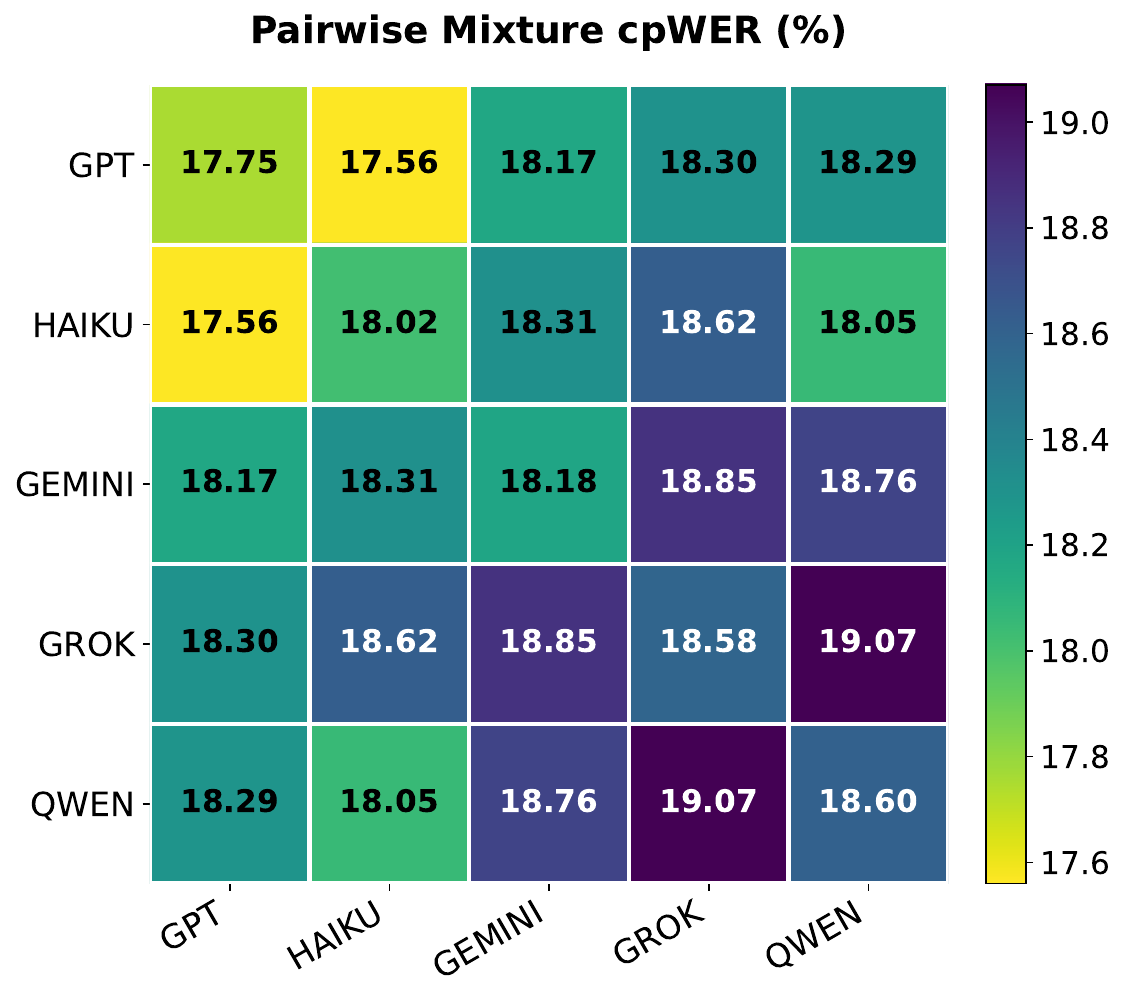}
    \end{minipage}
    \hfill
    \begin{minipage}{0.49\textwidth}
        \centering
        \includegraphics[width=\linewidth]{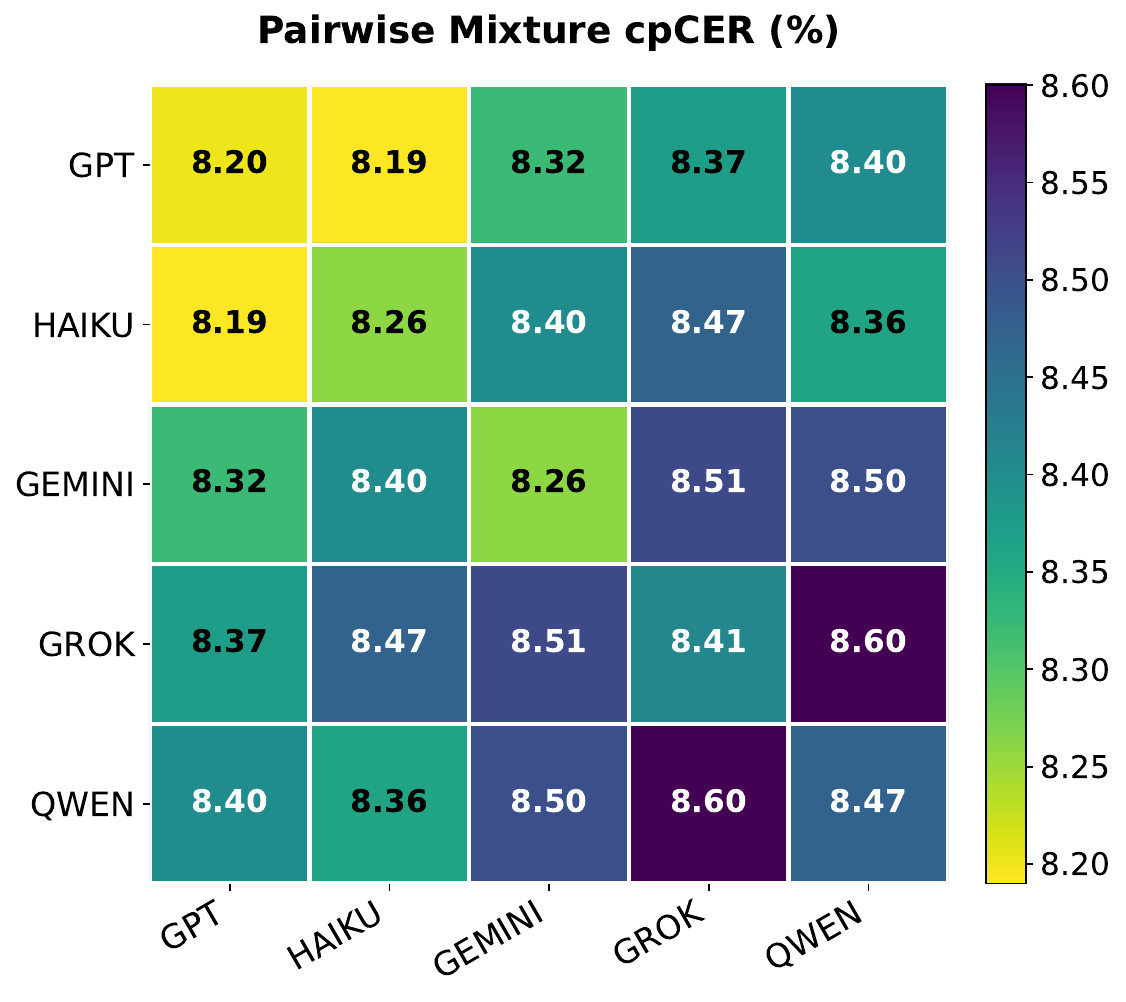}
    \end{minipage}

    \caption{Pairwise mixture performance on the BEA-Dialogue eval set. Diagonal entries correspond to single-generator configurations, while off-diagonal entries correspond to uniform two-generator mixtures.}
    \label{fig:mixture_heatmaps}
\end{figure*}

We next evaluate whether combining generators provides complementary benefits at a fixed synthetic-data budget. Table~\ref{tab:mixture_summary} reports the best subset at each mixture cardinality, while Figures~\ref{fig:mixture_upset} and~\ref{fig:mixture_heatmaps} visualize the same pattern across subset and pairwise comparisons. The best fixed-budget mixture is GPT + Haiku, which reaches 8.19 cpCER and 17.56 cpWER. This improves over the best single-generator condition by 0.19 absolute cpWER and 0.01 absolute cpCER, showing that the strongest pair contributes is mildly complementary.

The improvement does not continue as additional generators are added under the same 500-conversation budget. The best three-generator subset has higher cpWER (17.87), and the best four- and five-generator subsets degrade further to 18.19 and 18.27. Thus, diversity across LLM families is not automatically beneficial: when the total number of synthetic conversations is fixed, adding weaker or less complementary generators can replace higher-value GPT/Haiku samples and reduce the average utility of the augmentation set. Naturally, reducing the proportion of GPT and Haiku also decreases the total duration of synthetic conversations. However, the order in which models are introduced to achieve optimal performance does not align with their ranking in the single-generator setting, indicating that scale alone is not the determining factor.

Figure~\ref{fig:mixture_upset} makes this non-monotonic subset behavior explicit. The highest-performing subsets are concentrated around combinations involving GPT and Haiku, whereas larger mixtures do not consistently outperform smaller ones despite incorporating a greater number of LLM families. Notably, Qwen appears in two of the top five configurations, while Gemini is absent from all of them, even though Gemini substantially outperformed Qwen in the single-generator setting shown in Table~\ref{tab:single_generator_best}. This observation suggests that the contribution of a generator to a mixture cannot be inferred solely from its standalone performance; rather, the interactions among generators play a critical role in determining the effectiveness of the resulting training data.

The pairwise heatmaps in Figure~\ref{fig:mixture_heatmaps} provide a complementary view of the same result. The diagonal entries show the single-generator baselines, and the off-diagonal entries show which pairings improve over their individual components. Together with the UpSet plot, the heatmaps indicate that complementarity is concentrated in a small number of pairings rather than being a general property of mixing any two LLMs.

\subsection{Scale-up and baseline comparison}
\begin{table*}[t]
\centering
\small
\resizebox{0.98\textwidth}{!}{%

\begin{tabular}{lp{6.0cm}p{3.0cm}cc}
\toprule
\textbf{Setup} & \textbf{Configuration} & \textbf{Train size (h)} & \textbf{cpCER} & \textbf{cpWER} \\
\midrule
Baseline 1 & Whisper (z-s) & N/A & 12.18 & 22.13 \\
Baseline 2 & 2700h monolingual (z-s) & 2700 & \underline{7.71} & \underline{16.27} \\
\midrule
Baseline 3 & BEA-Dialogue only & 67 & 9.00 & 20.44 \\
Baseline 4 & BEA-Dialogue + simulated & 67 + 209 & \textit{8.13} & \textit{17.64}  \\
\midrule
1-scale & GPT & 67 + 146 & 8.20 & 17.75 \\
2-scale & GPT + Haiku & 67 + 273 & 8.03 & 16.96 \\
3-scale & GPT + Haiku + Qwen & 67 + 346 & 8.05 & 16.86  \\
\textbf{4-scale} & \textbf{GPT + Haiku + Qwen + Grok} & 67 + 427 & \textbf{7.97} & \textbf{16.65}  \\
5-scale & GPT + Haiku + Qwen + Grok + Gemini & 67 + 512 & 8.06 & 16.68 \\
\midrule
4-scale + sim & GPT + Haiku + Qwen + Grok + simulated & 67 + 427 + 209  & \underline{\textbf{7.57}} & \underline{\textbf{15.40}} \\
\bottomrule
\end{tabular}%
}
\caption{Baseline and scale-up comparison on the held-out BEA-Dialogue eval set. Scale-up rows add the full available synthetic set from each listed generator, while the final row also adds simulated BEA-Large-derived conversations. Abbreviation: z-s = zero-shot.}
\label{tab:scaleup_summary}
\end{table*}

Finally, we test whether the generator combinations identified above remain useful when the fixed 500-conversation budget is removed. In this setting, each scale-up condition uses the full available synthetic set from the selected generators, so adding a generator increases the total amount of synthetic speech rather than redistributing a constant budget. Table~\ref{tab:scaleup_summary} compares these scale-up runs with the baselines introduced in Section~4.3.

The scale-up results show that additional synthetic data is most effective when it comes from complementary, high-utility generators. GPT alone reaches 17.75 cpWER with 146 synthetic hours. Adding Haiku produces the largest marginal gain, reducing cpWER to 16.96 and cpCER to 8.03 with 273 synthetic hours. Adding Qwen yields a smaller word-level improvement (16.86 cpWER), while character-level accuracy degrades slightly relative to the two-generator condition. The best purely LLM-generated scale-up system is the four-generator GPT + Haiku + Qwen + Grok configuration, which reaches 7.97 cpCER and 16.65 cpWER with 427 synthetic hours. In contrast, adding Gemini increases the synthetic pool to 512 hours but slightly worsens both metrics, indicating that scale alone is insufficient: the marginal generator must contribute useful acoustic, lexical, or conversational variation rather than simply increasing the amount of training audio.

The comparison with baselines reinforces this interpretation. The four-generator scale-up model reduces cpWER by 3.79 points and cpCER by 1.03 points relative to BEA-Dialogue-only training, and it also outperforms the speaker-aware simulated-conversation baseline. Despite using only 67 hours of real conversational speech plus synthetic augmentation, it approaches the 2700-hour Hungarian-trained zero-shot system. This suggests that scenario-driven synthetic conversations that match the style of the target corpus can provide more relevant training context to the model than large-scale generic training, which focuses more on monologue speech and broadcast data.

The final row combines the best LLM scale-up condition with the reproduced BEA-Large-derived simulated conversations of \citet{Gedeon2026SASC}. This combined system further reduces cpWER from 16.65 to 15.40 and cpCER from 7.97 to 7.57, yielding the best result in Table~\ref{tab:scaleup_summary}. The gain over the pure LLM scale-up condition shows that the two augmentation sources are complementary: LLM-generated conversations add controlled topic and dialogue diversity, while simulations from real utterances preserve natural acoustic and lexical properties. The combined model also surpasses the Hungarian-trained zero-shot baseline while using only a fraction of the 2700 hours as a training set, most of which was synthetic. Given the human effort required to annotate the 2700 hours of speech, this result suggests a more efficient use of resources while preserving strong recognition performance.

To assess statistical significance, we apply the bootstrap-based significance testing procedure of \citet{Confidence_Intervals} to the evaluation set using a significance level of $\alpha = 0.05$. The results are summarized as follows:
\begin{itemize}
\item Baselines 1 and 3 are significantly outperformed by all scaled models with respect to both cpCER and cpWER.
\item Baseline 4 is significantly outperformed by the 2-, 3-, 4-, and 5-scale models in terms of cpWER. For cpCER, however, a statistically significant improvement is observed only for the \textit{4-scale + sim} model; all other comparisons are not statistically significant.
\item Relative to Baseline 2, the \textit{4-scale + sim} model achieves a statistically significant improvement in cpWER, whereas no significant difference is observed in cpCER.
\end{itemize}

The results support four conclusions. First, LLM-generated conversational augmentation improves Hungarian conversational ASR even when a single generator is used. Second, generator mixtures help selectively: GPT and Haiku are complementary, but larger fixed-budget mixtures are weaker because they dilute the strongest sources. Third, when the fixed-budget constraint is removed, scaling the amount of synthetic speech can yield further gains, provided that the added generators contribute useful variation. Fourth, combining TTS-based synthetic conversations with simulated conversations derived from real utterances provides the strongest evaluated system, showing that generated and real-utterance-based augmentation capture complementary aspects of conversational speech.

\section{Conclusion}
We presented an LLM-driven augmentation pipeline for conversational ASR that combines scenario-level dialogue generation, metadata-conditioned TTS synthesis, and speaker-aware conversation simulation. By generating synthetic conversations with explicit participant attributes and converting them into multi-speaker training audio, the approach expands both the linguistic and conversational diversity available to the ASR model. Across the experiments in Hungarian with several evaluated LLM families, synthetic augmentation consistently improves over BEA-Dialogue-only training, with the strongest gains obtained when high-performing generators are scaled and combined with simulated BEA-Large-derived conversations.

The results show that generator choice and data composition are critical. Single-generator augmentation is already beneficial, but fixed-budget mixtures help only when the added generators are complementary; larger mixtures can degrade performance when they dilute stronger sources. When the augmentation budget is allowed to grow, however, scaling the strongest generators yields further improvements, and the final combined system achieves the best result in our experiments, surpassing a zero-shot monolingual Hungarian model trained on 2700 hours of speech. These findings suggest that LLM-generated conversational data is a practical complement to real conversational corpora and large-scale acoustic pretraining, especially in settings where domain-matched labeled speech is limited.

Our work opens several avenues for future research. First, scaling the experiments beyond the 500 conversations used in this study may further improve performance and provide a more comprehensive assessment of the proposed approach. Second, evaluating the pipeline on more domain-specific corpora could better demonstrate its strengths and reveal its effectiveness in specialized application scenarios. Such investigations would contribute to a deeper understanding of the pipeline's scalability, robustness, and domain adaptability.

\section*{Acknowledgment}
Project No. 2025-2.1.2-EKÖP-KDP-2025-00005 has been implemented with the support provided by the Ministry of Culture and Innovation of Hungary from the National Research, Development and Innovation Fund, financed under the EKÖP\_KDP-25-1-BME-21 funding scheme.

The work was also partially supported by the Hungarian NRDI Fund through the projects NKFIH K143075 and K135038, NKFIH-828- 2/2021(MILAB).

\newpage
\bibliography{tacl2021}
\bibliographystyle{acl_natbib}



\end{document}